\documentclass{interact}

\usepackage{titlesec}
\usepackage{graphicx}
\usepackage[version=3]{mhchem} 
\usepackage{graphicx}
\usepackage{subcaption}
\usepackage{array,multirow}
\usepackage{amsmath}
\usepackage{todonotes}
\usepackage{hyperref}
\usepackage{soul}
\usepackage{algpseudocode,algorithm,algorithmicx}
\usepackage[utf8x]{inputenc}

\usepackage{fancyhdr}

\begin{document}

\title{Embeddings as representation for symbolic music.}

\author{
\name{Sebastian Garcia-Valencia\textsuperscript{a,b}\thanks{Correspondence: Sebastian Garcia Valencia, Computer Science Department, Universidad EAFIT Carrera 49 No 7 Sur-50 Medellin, Colombia. Email: sgarci18@eafit.edu.co}}
\affil{\textsuperscript{a}Computer Science Department, Universidad EAFIT, Medellin, Colombia; \textsuperscript{b}Research and Development Department, AIVA Technologies, Luxembourg City, Luxembourg}
}

\maketitle

\begin{abstract}
A representation technique that allows encoding music in a way that contains musical meaning would improve the results of any model trained for computer music tasks like generation of melodies and harmonies of better quality. The field of natural language processing has done a lot of work in finding a way to capture the semantic meaning of words and sentences, and word embeddings have successfully shown the capabilities for such a task. In this paper, we experiment with embeddings to represent musical notes from 3 different variations of a dataset and analyze if the model can capture useful musical patterns. To do this, the resulting embeddings are visualized in projections using the t-SNE technique.
\end{abstract}


\section{Introduction} \label{sec:introduction}

The breakthroughs seen in last years in machine learning have attracted a lot of attention into the field, usually, a lot of effort is put into the models which correspond to the training phase of the project, however, the way the data is represented is not a trivial aspect and it can easily improve the results of the models without having to do any change in them.

Particularly in Natural Language Processing (NLP), the necessity of a good representation of the words that achieves some implicit context understanding is important \cite{Turian2010}. A typical representation to feed in a machine learning model is the binary one-hot vector, in this case, an array with as many positions as words in the vocabulary is created, and the words are represented by a version of the array containing a one digit in the position corresponding to the word. For example, the sentence "I like eating bread and eating cheese", would have as vocabulary the set {"I", "like", "eating", "bread", "and", "cheese"}, thus the representation of this words would be 6-dimensional binary one-hot vectors like "I" = 100000, "like" = 010000, cheese = 000001. As you can imagine, this representation has no context understanding at all, since all words are completely independent,  "bread" and "cheese" are as different between them as "I" and "like", which for a human is not the case.

We can make a similar exercise with music and imagine a vocabulary were instead of words we have notes, similarly to NLP we would like to include in the representation some understanding of musically related notes.

There is a common representation technique used in NLP called `word embedding' \cite{DBLP:journals/corr/abs-1301-3781} which has successfully solved this problem, it transforms words in points in a vectorial space. Embeddings can locate semantically close words in near spacial segments. The context can be interpreted using the words surrounding the target in the sentences or go further using the syntactic context \cite{levy2014}. Authors in \cite{baroni-dinu-kruszewski:2014:P14-1} shows how the approach of predicting the context instead of just counting words yields superior results.

Numerous ways to represent music has been used across the years in different research, the most predominant method is the use of representations based in arrays, which can be direct representations like in the 12-bit codification in \cite{Hornel1998} or the binary array used in \cite{Johnson2017}. Another option is making dimensional transformations of the array like \cite{Kaliakatsos-Papakostas2010} with its Dodecaphonic Trace Vector and \cite{DePrisco2017} and its token.
    
Word embeddings \cite{DBLP:journals/corr/abs-1301-3781} fall in the latter category and has been already successfully used to represent music. In \cite{sephora2016, herremans2017} the authors used it as a way to extract important patterns.

\subsection{t-SNE algoritm}
An important aspect with embeddings is the visualization, as the vectorial spaces where the points are located are high dimensional, we need some technique to make projections in 2 and 3 dimensions so that we can have a visual insight of the patterns being captured. t-Distributed Stochastic Neighbor Embedding (t-SNE) \cite{VanDerMaaten2008} is a non-linear dimensionality reduction technique where the similarity in the low dimensional space (projection) is calculated based in conditional probabilities. Its non-linearity lets the method preserve the local structure of the data, which makes it good projecting clusters.
\\

In this experiment we use a model thought originally for words \cite{DBLP:journals/corr/MikolovSCCD13} with a monophonic music dataset and show the patterns and clusters that emerge using the t-SNE visualization algorithm with tensorboard, we also analyzed how the embeddings vary respect to the type of database (normal, augmented through transposition, and represented as intervals instead of notes). Section \ref{sec:results} analyse the resulting embeddings. 

The remaining part of this paper is organized as follows: Chapter \ref{sec:experiments} presents the dataset and procedure done. Chapter \ref{sec:results} analyses the patterns found in the projections and tables. Finally, chapter \ref{sec:conlusions}, give general conclusions.

\section{Experiments} \label{sec:experiments}
\subsection{Dataset}
We use the mono-midi-transposition-dataset, this is a dataset containing 3 variations of monophonic pieces from musescore: 1) The control dataset where each element xi in the x array is each note of each piece and each element yi in the y array is the subsequent note of element xi. 2) The DB12 dataset which is a data augmentation of the control dataset created making a transposition of each piece in it to the other 11 tonalities, making this dataset 12 times bigger. 3) The interval dataset, where we take the relative semitone changes between notes instead of the notes itself i.e. (C4,D4,G4,F4) $\xrightarrow{}$ (2,5,-2).

\subsection{Procedure}
A recurrent neural network based in LSTM cells was trained using the 3 datasets, in the first part of the neural network an embedding component containing an encoder transforms the elements of the sequence in embeddings with a dimension of 128, this means each note or interval will be represented as a point and positioned in a vectorial space with 128 dimensions, the embeddings are trained simultaneously with the neural network, and at the end, projections in 2D and 3D are created using the t-SNE algorithm.

\section{Results} \label{sec:results}

\begin{figure}[h]
    \centering
    \includegraphics[width=0.4\textwidth]{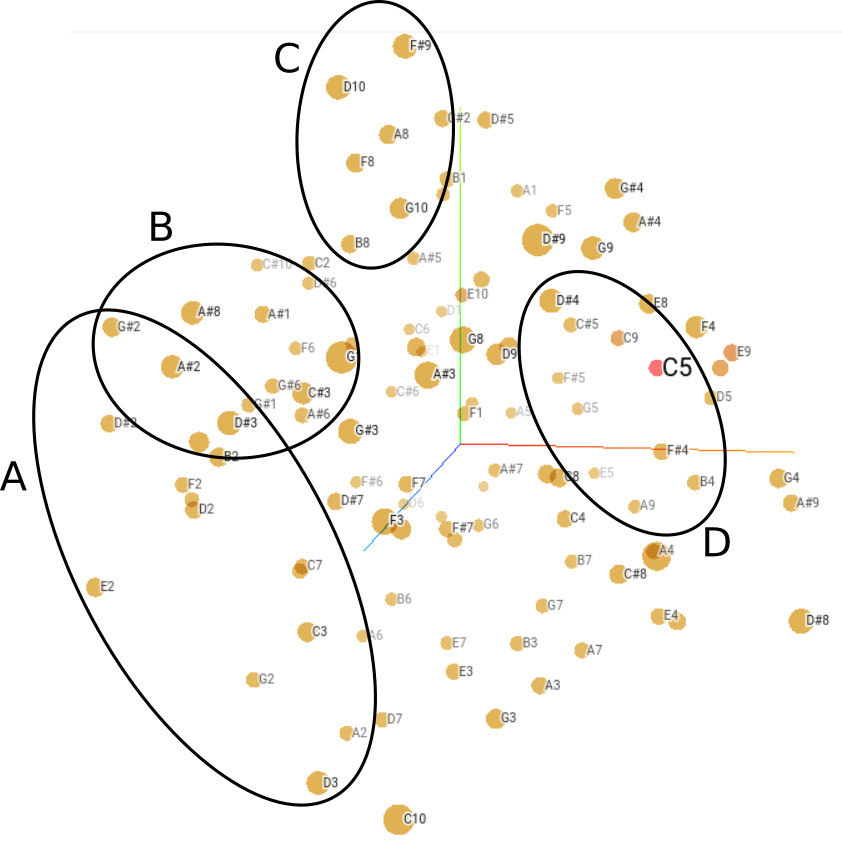}
    \caption{Control dataset 3D projection of Embeddings (size 128)}
    \label{fig:embeddingscontrollstm}
\end{figure}

To understand which patterns are possible to learn with the embeddings, this section visualizes one sample of the embeddings and analyses some concrete cases for each dataset. Figure \ref{fig:embeddingscontrollstm} shows a projection in 3d of the embeddings which have an original dimension of 128, for the selected model for the control dataset and LSTM cell. We can visually notice some groups: very low notes from octaves 2 and 3 (fig. \ref{fig:embeddingscontrollstm}.A), high notes from octaves 8, 9 and 10 (fig. \ref{fig:embeddingscontrollstm}.C), middle notes from the most common octaves 4 and 5 (fig. \ref{fig:embeddingscontrollstm}.D) and finally, notes from very different octaves, but all with alterations (fig. \ref{fig:embeddingscontrollstm}.B). This exemplifies the kind of semantic patterns that embeddings allow to catch.

\begin{figure}[h!]
    \centering
    \begin{subfigure}[b]{0.4\textwidth}
        \includegraphics[width=\textwidth]{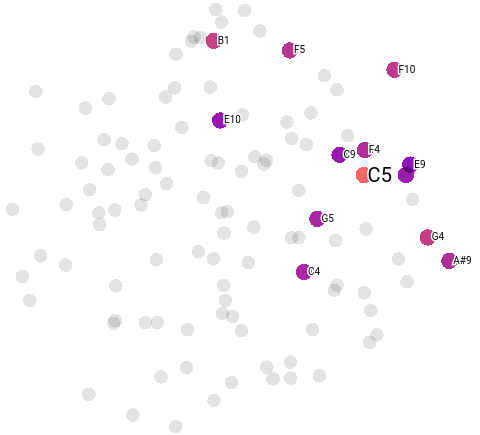}
        \caption{Control datasaet}
        \label{fig:control_neighbours}
    \end{subfigure} 
    \begin{subfigure}[b]{0.4\textwidth}
        \includegraphics[width=\textwidth]{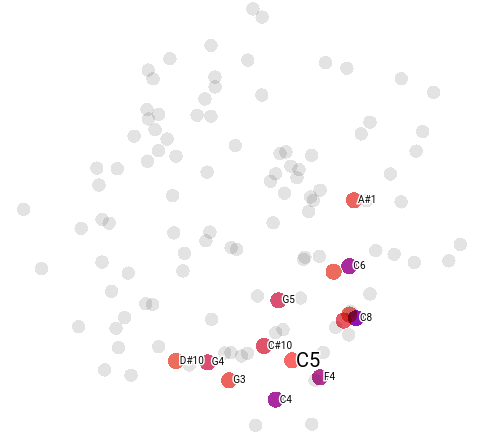}
        \caption{DB12 datasaet}
        \label{fig:db12_neighbours}
    \end{subfigure}
    \begin{subfigure}[b]{0.4\textwidth}
        \includegraphics[width=\textwidth]{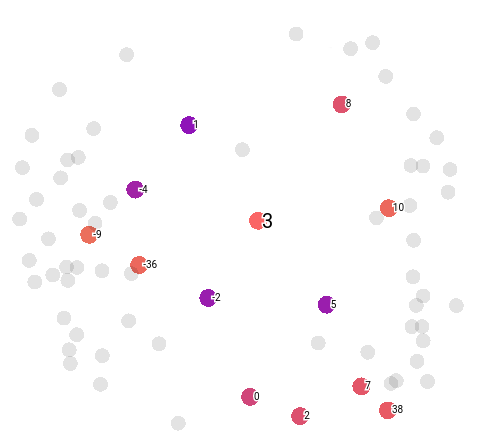}
        \caption{Interval datasaet}
        \label{fig:interval_neighbours}
    \end{subfigure}     
 
    \caption{2D Projection of 10 nearest neighbours to selected points}\label{fig:interval}
\end{figure}

Table \ref{tab:nearest10pointsembeddings} and figures \ref{fig:control_neighbours}, \ref{fig:db12_neighbours} and \ref{fig:interval_neighbours} summarises some information of the embeddings and shows the 10 nearest neighbours to a selected point for each dataset. In the case of the control and DB12 versions, the embeddings correspond to notes. It is interesting that while the control case has 111 embeddings, the DB12 (12 times bigger) only has 118, the reason is that even if the DB12 has 12 times more songs, the notes these songs have will be in the same possible range of midi notes (0-127). 

For the embeddings of these 2 datasets, the selected point was central C (C5)\footnote{Here is important to clarify that even when the acoustical society of America (ASA) defines the central C as the one in the fourth octave (C4), there are other conventions for the number accompanying the note name. This is the case for the most midi hardware manufacturers who take the octave 0 (corresponding to the midi notes 0-11) as the lowest octave, in opposition, to ASA that defines octave -2 as the lowest. Therefore, the midi table assigns 60 (central C) to C5. Just keep in mind, that every time that central or middle C appears in the paper, it refers to the 261 Hz C.}, in the control case, the nearest neighbors are from quite distant octaves (E10, E9 and C9) and from the same octave are the notes F and G (tab. \ref{tab:embeddingbaselstm} and fig \ref{fig:control_neighbours}). 

For DB12 (tab. \ref{tab:embeddingdb12lstm} and fig \ref{fig:db12_neighbours}), the 3 nearest points in the vectorial space are the same note in different octaves (C4, C6 and C8). From the same octave are the notes F and G. The relation with C through the different octaves can be the consequence of a better understanding of the note because of the dataset augmentation.

In the case of the interval dataset, the embeddings represent the changes between notes; there are 76 of them in the data set (tab. \ref{tab:embeddingintervallstm} and fig \ref{fig:interval_neighbours}). In this case, the point selected was positive 3, this represents a change of 3 semitones up (a minor third), which is a typical interval in music. The nearest points are a minor second (1), a major second below (-2), a perfect fourth (5), a major third below (-4) and a unison (0), which means a repetition of the same note.

\begin{table}[h!]
\begin{subtable}[t]{.3\linewidth}
\centering
\begin{tabular}{cl} \hline
\hline
Total:&111\\
Selection:&C5 \\ \hline
\hline
Note & Cos   \\ \hline 

E9  & 0.505 \\
E10  & 0.526 \\
C9  & 0.531 \\
B9 & 0.546 \\
G5 & 0.590 \\
C4 & 0.599 \\ 
A♯9 & 0.630 \\
F4 & 0.637 \\
F5 & 0.669 \\
F10 & 0.697 \\
\hline
\hline

\end{tabular}

\caption{Control Lstm}  \label{tab:embeddingbaselstm}
\end{subtable}
\begin{subtable}[t]{.3\linewidth}
\centering
\begin{tabular}{cl} \hline
\hline
Total:& 118 \\
Selection:& C5 \\ \hline
\hline
Note & Cos   \\ \hline 

C8 & 0.550\\
C4 & 0.661\\
C6 & 0.603\\
F4 & 0.630\\
G4 & 0.722\\
G5 & 0.725\\
C♯10 & 0.743\\
F5 & 0.764\\
G3 & 0.804\\
A♯1 & 0.806\\
\hline
\hline

\end{tabular}

\caption{DB12 LSTM}\label{tab:embeddingdb12lstm}
\end{subtable}
\begin{subtable}[t]{.3\linewidth}
\centering
\begin{tabular}{cl} \hline
\hline
Total:& 76 \\
Selection:& 3 \\ \hline
\hline
Interval & Cos   \\ \hline 

1 & 0.473 \\
-2 & 0.489 \\
5 & 0.491 \\
-4 & 0.502 \\
0 & 0.595 \\
2 & 0.626 \\
8 & 0.632 \\
7 & 0.645 \\
38 & 0.658 \\
10 & 0.711 \\
\hline
\hline

\end{tabular}

\caption{Interval LSTM}\label{tab:embeddingintervallstm}
\end{subtable}

\caption{Nearest 10 neighbours to selected points} \label{tab:nearest10pointsembeddings}
\end{table}

\section{Conclusions} \label{sec:conlusions}

The results of section \ref{sec:results} show that embeddings can capture semantic patterns in the musical context exactly as they do with words in NLP. No matter if the data used is based on notes like is the case with the control and DB12 dataset, or intervals, the vectorial space where the points are represented can capture features as octave relationship between pitch classes as shows figure \ref{fig:db12_neighbours} and table \ref{tab:embeddingdb12lstm}, or notes with alteration clusters separated to natural notes as shown in figure \ref{fig:embeddingscontrollstm}.B. It is also remarkable how the closest neighbors to the minor third interval in the case of the interval dataset are other members of the 12 main intervals like the minor second, the perfect fourth and the major third.

The potential of encoding the data as embeddings before feeding it into a predictive model like a neural network to improve the learning results is promising since this way the data is going to have already some implicit knowledge from scratch.

\bibliographystyle{apalike}
\bibliography{bibliography.bib}

\begin{thebibliography}{}

\bibitem[Baroni et~al., 2014]{baroni-dinu-kruszewski:2014:P14-1}
Baroni, M., Dinu, G., and Kruszewski, G. (2014).
\newblock {Don't count, predict! A systematic comparison of context-counting
  vs. context-predicting semantic vectors}.
\newblock In {\em Proceedings of the 52nd Annual Meeting of the Association for
  Computational Linguistics (Volume 1: Long Papers)}, pages 238--247,
  Baltimore, Maryland. Association for Computational Linguistics.

\bibitem[{De Prisco} et~al., 2017]{DePrisco2017}
{De Prisco}, R., Malandrino, D., Zaccagnino, G., Zaccagnino, R., and Zizza, R.
  (2017).
\newblock {A Kind of Bio-inspired Learning of mUsic stylE}.
\newblock In Correia, J., Ciesielski, V., and Liapis, A., editors, {\em
  Computational Intelligence in Music, Sound, Art and Design: 6th International
  Conference, EvoMUSART 2017, Amsterdam, The Netherlands, April 19--21, 2017,
  Proceedings}, pages 97--113. Springer International Publishing, Cham.

\bibitem[Herremans and Chuan, 2017]{herremans2017}
Herremans, D. and Chuan, C.-H. (2017).
\newblock {Modeling Musical Context with Word2vec}.
\newblock {\em CoRR}, abs/1706.0.

\bibitem[H{\"{o}}rnel, 1998]{Hornel1998}
H{\"{o}}rnel, D. (1998).
\newblock {MELONET I: Neural nets for inventing baroque-style chorale
  variations}.
\newblock In {\em Advances in Neural Information Processing Systems}.

\bibitem[Johnson, 2017]{Johnson2017}
Johnson, D.~D. (2017).
\newblock {Generating Polyphonic Music Using Tied Parallel Networks}.
\newblock In Correia, J., Ciesielski, V., and Liapis, A., editors, {\em
  Computational Intelligence in Music, Sound, Art and Design: 6th International
  Conference, EvoMUSART 2017, Amsterdam, The Netherlands, April 19--21, 2017,
  Proceedings}, pages 128--143. Springer International Publishing, Cham.

\bibitem[Kaliakatsos-Papakostas et~al., 2010]{Kaliakatsos-Papakostas2010}
Kaliakatsos-Papakostas, M.~A., Epitropakis, M.~G., and Vrahatis, M.~N. (2010).
\newblock {Musical Composer Identification through Probabilistic and
  Feedforward Neural Networks}.
\newblock In {Di Chio}, C., Brabazon, A., {Di Caro}, G.~A., Ebner, M., Farooq,
  M., Fink, A., Grahl, J., Greenfield, G., Machado, P., O'Neill, M., Tarantino,
  E., and Urquhart, N., editors, {\em Applications of Evolutionary Computation:
  EvoApplications 2010: EvoCOMNET, EvoENVIRONMENT, EvoFIN, EvoMUSART, and
  EvoTRANSLOG, Istanbul, Turkey, April 7-9, 2010, Proceedings, Part II}, pages
  411--420. Springer Berlin Heidelberg, Berlin, Heidelberg.

\bibitem[Levy and Goldberg, 2014]{levy2014}
Levy, O. and Goldberg, Y. (2014).
\newblock {Dependency-Based Word Embeddings}.
\newblock In {\em 52nd Annual Meeting of the Association for Computational
  Linguistics, ACL 2014 - Proceedings of the Conference}, volume~2, pages
  302--308.

\bibitem[Madjiheurem et~al., 2016]{sephora2016}
Madjiheurem, S., Qu, L., and Walder, C. (2016).
\newblock {Chord2Vec: Learning Musical Chord Embeddings}.

\bibitem[Mikolov et~al., 2013a]{DBLP:journals/corr/abs-1301-3781}
Mikolov, T., Chen, K., Corrado, G., and Dean, J. (2013a).
\newblock {Efficient Estimation of Word Representations in Vector Space}.
\newblock {\em CoRR}, abs/1301.3.

\bibitem[Mikolov et~al., 2013b]{DBLP:journals/corr/MikolovSCCD13}
Mikolov, T., Sutskever, I., Chen, K., Corrado, G., and Dean, J. (2013b).
\newblock {Distributed Representations of Words and Phrases and their
  Compositionality}.
\newblock {\em CoRR}, abs/1310.4.

\bibitem[Turian et~al., 2010]{Turian2010}
Turian, J., Ratinov, L., and Bengio, Y. (2010).
\newblock {Word representations: A simple and general method for
  semi-supervised learning}.
\newblock pages 384--394.

\bibitem[{Van Der Maaten} and Hinton, 2008]{VanDerMaaten2008}
{Van Der Maaten}, L. and Hinton, G. (2008).
\newblock {Visualizing Data using t-SNE}.
\newblock Technical report.

\end{thebibliography}
\end{document}